\documentclass[conference]{IEEEtran}
\IEEEoverridecommandlockouts
\usepackage{cite}
\usepackage{amsmath,amssymb,amsfonts}
\usepackage{algorithmic}
\usepackage{graphicx}
\usepackage{textcomp}
\usepackage{xcolor}
\usepackage[font=small,skip=0pt]{caption}
\usepackage[linesnumbered,ruled,vlined]{algorithm2e}
\usepackage{times}
\usepackage{epsfig}
\usepackage{graphicx}
\usepackage{amsmath}
\usepackage{amssymb}
\usepackage{relsize}
\usepackage{url}
\SetKwInput{KwInput}{Input}                % Set the Input
\SetKwInput{KwOutput}{Output}              % set the Output

\SetCommentSty{mycommfont}
\usepackage{subcaption}
\usepackage{xurl} % to linebreak url anywhere (in bibliography)
\usepackage{url}
\usepackage[hidelinks]{hyperref}
\usepackage{booktabs}
\usepackage{multirow}
\usepackage{tabularx}
\def\BibTeX{{\rm B\kern-.05em{\sc i\kern-.025em b}\kern-.08em
    T\kern-.1667em\lower.7ex\hbox{E}\kern-.125emX}}
\newcommand{\myK}{{\textnormal{\fontfamily{cmr}\selectfont K}}}
\begin{document}
\pagestyle{plain}
\pagenumbering{arabic}
%%%%%%%%% TITLE
\title{GenConViT: Deepfake Video Detection Using Generative Convolutional Vision Transformer}

\author{\IEEEauthorblockN{Deressa Wodajo Deressa\textsuperscript{*},
Hannes Mareen\textsuperscript{**}, 
Peter Lambert\textsuperscript{**}, \\
Solomon Atnafu\textsuperscript{\dag}, 
Zahid Akhtar\textsuperscript{\ddag},
Glenn Van Wallendael\textsuperscript{**}}
\IEEEauthorblockA{\textsuperscript{**}Ghent University -- imec, IDLab, Department of Electronics and Information Systems, Gent, Belgium\\
firstname.lastname@ugent.be, \url{https://media.idlab.ugent.be}}
\IEEEauthorblockA{\textsuperscript{*}deressawodajo.deressa@ugent.be}
\IEEEauthorblockA{\textsuperscript{\dag}Addis Ababa University, \url{solomon.atnafu@aau.edu.et}}
\IEEEauthorblockA{\textsuperscript{\ddag}State University of New York Polytechnic Institute, USA, \url{akhtarz@sunypoly.edu}}
}

\maketitle
%%%%%%%%% ABSTRACT
\begin{abstract}
   Deepfakes have raised significant concerns due to their 
   potential to spread false information and compromise 
   digital media integrity. Current deepfake detection models often struggle to generalize across a diverse range of deepfake generation techniques and video content. In this work, we propose a 
   Generative Convolutional Vision Transformer (GenConViT) 
   for deepfake video detection. Our model combines
   ConvNeXt and Swin Transformer models for feature 
   extraction, and it utilizes Autoencoder and Variational 
   Autoencoder to learn from the latent data distribution. 
   By learning from the visual artifacts and latent data distribution, 
   GenConViT achieves improved performance in detecting a 
   wide range of deepfake videos. The model is trained and 
   evaluated on DFDC, FF++, TM, DeepfakeTIMIT, and Celeb-DF (v$2$) 
   datasets. The proposed GenConViT model 
   demonstrates strong performance in deepfake video 
   detection, achieving high accuracy across the tested datasets. 
   While our model shows promising results in deepfake video detection by leveraging visual and latent features, we demonstrate that further work is needed to improve its generalizability, i.e., when encountering out-of-distribution data. Our model provides an effective
   solution for identifying a wide range of fake videos while 
   preserving media integrity.
   The open-source code for GenConViT is available at \url{https://github.com/erprogs/GenConViT}.

\end{abstract}

%%%%%%%%% BODY TEXT
\section{Introduction}
Deepfakes are hyper-realistic manipulated media generated using various advanced Deep Learning (DL) techniques. Deepfake videos are produced by superimposing one person's facial features onto another person's face through techniques such as face replacement, facial reenactment, face editing, and complete face synthesis~\cite{mnmjim23,tolosana_introduction_2022}, among others. Recently, deepfakes have become a public concern due to their potential for misuse and the spread of false information~\cite{dv22, jwhgml22, z20}. Various deepfake creation methods are readily available for use, and anyone interested can use them to easily modify existing media, presenting a false representation of reality where individuals appear to be speaking or performing actions that never actually took place. Additionally, the manipulation of the videos is done on a frame-by-frame basis, making deepfakes seem more realistic and believable. Consequently, deepfake creation techniques have been used to manipulate images and videos, causing celebrities and politicians to be depicted as saying or doing things that are untrue.

The applications of deepfake videos can range from creative uses such as creating realistic special effects or replacing an actor in the film and entertainment industry, to potentially harmful attacks, such as using deepfakes to create misleading videos for criminal purposes~\cite{nnnnhnnpn22,Korshunov2022}. In the realm of politics, deepfakes have been utilized to create fake political videos, war propaganda, and videos with the intention of influencing elections. This has led to widespread concern about the potential for malicious actors to use deepfakes to spread false information, erode the credibility of political candidates, or even pose a threat to national security. The dissemination of misleading and false information through deepfakes can have real-life consequences, making it imperative to address the need for accurate and reliable deepfake video detection techniques.

Deepfake video detection attempts to detect if a given video has been tampered with or manipulated using deepfake techniques. The challenge of detecting deepfakes has inspired researchers and technology companies to develop various deep-learning methods to identify tampered videos~\cite{vk22,kak23,dv22,mnmjim23,llhwzzgl,urrehman}. Currently, deepfake detection methods often rely heavily on visual features~\cite{anye18,yll19,tpd22,mrs19}. The visual features-based detection can be on a frame-per-frame basis~\cite{hyk23} or temporal relationship between differences between frames through time~\cite{gcydlhm,kd21}. However, these methods mostly encounter challenges in detecting deepfakes that differ from the training data, thus, failing sufficient detection. In recent years, the rapid progress in advanced generative models like Generative Adversarial Networks (GAN)~\cite{gpmxwocb,klt,mo}, Variational Autoencoders (VAE)~\cite{kw,kw19}, and Diffusion Models (DM)~\cite{dn} has made it even more challenging to detect deepfakes based on visual artifacts alone since the deepfakes leave small to no trace of visual clues that help to distinguish them from the images from the real world. Several authors have proposed alternative detection methods, such as utilizing biological signals~\cite{cd}, geometric features~\cite{ysldzwl}, frequency information~\cite{wraq}, spatial features with temporal information~\cite{zlhwzg,kk22} and generative adversarial network fingerprints~\cite{ysaf}, as potential solutions to the difficulties of detecting deepfakes. Another approach to spotting deepfake videos involves examining the consistency of pixels or groups of similar pixels~\cite{gcsdv}.
Current deepfake video detection methods have a competitive result in identifying manipulated videos, but they often fail to generalize in more diverse videos, particularly in environments with different facial poses, light angles, and movements~\cite{li2024generalizabledeepfakeimagedetectors}.

In an effort to provide more generalization, we propose a novel architecture called the Generative Convolutional Vision Transformer (GenConViT), aimed at detecting a diverse set of fake videos. Our proposed architecture leverages both visual artifacts and latent data distribution of the data in its detection process. 

GenConViT has two main components: the generative part and the feature extraction part.
The generative part utilizes two different models (an Autoencoder (AE)~\cite{bkg21} and VAE), to learn the latent data distribution of the training data.
The feature extraction component employs ConvNext~\cite{lmwfds} and the Swin Transformer~\cite{llchwzlg} to extract relevant visual features.
Our proposed model addresses the limitations of current detection methods by learning both the visual artifacts and the latent data distribution, making it capable of detecting a wider range of fake videos. Extensive experimental results demonstrate that GenConViT has competitive results compared to the other state-of-the-art deepfake detection models. 

In summary, our paper has the following contributions:
\begin{itemize}
	\item GenConViT: We propose Generative Convolution Vision Transformer for deepfake video detection. Our method utilizes AE and VAE to generate images, extract the visual features of the original and generated image using ConvNext and the Swin Transformer, and then use the extracted features to detect whether the video is a deepfake video.
	\item By training our model on more than 1 million images extracted from five large deepfake dataset and detecting both visual and latent features, our proposed method aims to enhance  deepfake detection performance compared to previously proposed methods. 
    \item We thoroughly evaluated GenConViT on more than $3,972$ videos collected from five deepfake video datasets, demonstrating its robust performance under diverse experimental settings.
    \item We open-source the code and weights at \url{https://github.com/erprogs/GenConViT}.
    \item While our approach achieves strong performance, our comprehensive ablation study (leave-one-dataset-out) indicates a potential limitation in the generalizability of deepfake detection methods, particularly when faced with out-of-distribution data. This highlights an area for future work.
    
\end{itemize}
The remainder of this article is structured as follows. An overview of the existing works is presented in Section~\ref{sec:related}. The proposed Generative Convolutional Vision Transformer (GenConViT) deepfake detection framework and datasets are detailed in Section~\ref{sec:proposed}. Experiments are discussed in Section~\ref{sec:evaluation}, and an ablation study is performed in Section~\ref{sec:ablation}. Finally, the conclusions are drawn in Section~\ref{sec:conclusion}.
%-------------------------------------------------------------------------
\section{Related Work}
\label{sec:related}
In recent years, deepfakes have become more realistic as well as becoming harder to detect. In this section, we discuss the various deep learning methods used to create (Section~\ref{sec:related-generation}) and detect (Section~\ref{sec:related-detection}) deepfakes.

\subsection{Deepfake Generation}
\label{sec:related-generation}
Deepfake videos are created by advanced DL methods, such as GANs, VAEs, and DMs. Deepfake was first introduced in 2017 on the social media platform Reddit by a user called Deepfakes to showcase a DL technique to generate Deepfake videos~\cite{nnnnhnnpn22}. The videos and the accompanying code garnered significant attention, and soon people started to explore other ways to create a hyper-realistic video. Some widely used DL techniques to create deepfakes include face synthesis, face reenactment, face replacement, identity swapping, attribute manipulation, and expression swap, to name a few.
GANs are a type of Generative Model (GM) that consists of two Neural Networks (NNs), a generator $G$, which takes in random noise and produces synthetic data, and the discriminator $D$, which determines whether the input sample is real or fake. Another NN used to create Deepfakes is VAE, which consists of an Encoder and a Decoder NN. The Encoder encodes the input image into a lower dimensional latent space, and the decoder reconstructs an image from the latent space. Convolutional Neural Networks (CNNs)~\cite{on,hj19} are also used for face synthesis by learning the mapping between a face image and its attributes, such as facial expression, age, and gender.

Face synthesis~\cite{kstoa22} is a method of synthesizing desired non-existent face images based on a given input image. In face synthesis, a generator $G$ generates a face image, and a discriminator $D$ discriminates whether the sample is real or fake. Face reenactment~\cite{tzstn,htw22,sdlsrs} is a face synthesis method that transfers source face attributes to a target face while preserving the appearance and identity of the target face’s features.
Face2Face~\cite{tzstn} is a real-time face reenactment technique that animates the facial expressions of the target video by using a source actor and generates manipulated output video. Some other examples of face synthesis methods include frontal view synthesis, changing the facial pose from an input image, altering facial attributes, aging the face to create diverse and realistic results. Face swapping~\cite{ksdt,lbycw,zlwxs} is the process of replacing a person's face in an image or video with the face of another person to create a non-existent realistic face. 

Several types of GANs have been proposed for face synthesis, including Progressive GANs~\cite{kall}, Wasserstein GANs~\cite{acb17}, and Style-Based GANs~\cite{klt}. Progressive GANs allow for the generation of high-resolution images by gradually increasing the resolution of the generated images throughout the training process. Wasserstein GANs improve the stability of the GAN training process by using a different loss function. Style-Based GANs allow for the control of certain style aspects of the generated faces, such as facial expression and hair style. Conditional GANs~\cite{mo}, also known as supervised GANs, use labeled data to generate facial semantics. Paired image-to-image translation GANs~\cite{izze,plqc} are a type of conditional GAN that translate an input image from one domain to another, given input-output pairs of images as training data. Pix2Pix is one example of a paired image-to-image translation GAN.

In conclusion, deepfake generation leverages a diverse range of deep learning methods, including GAN architectures (e.g., Progressive, Wasserstein, StyleGANs), Transformers, VAEs, and CNNs. These methods are applied to create hyper-realistic images and videos through techniques such as face synthesis, reenactment, and face swapping. This rapid advancement and the diversity of approaches pose a significant and evolving challenge for deepfake detection.

\subsection{Deepfake Detection}
\label{sec:related-detection}
A key question in the deepfake detection pipeline is determining whether a given video is fake or real. With the advancement of DL methods that can create hyperrealistic images, deepfake detection has increasingly become a challenging task. To this effect, various authors have proposed deepfake detection techniques that use different approaches, including visual features, biological signals, and frequency information, to name a few\cite{Gragnaniello2022, hmareencompress}.

Several deepfake detection methods rely on extracting visual features from manipulated videos. MesoNet~\cite{anye18} is a deepfake detection method that uses CNNs to extract mesoscopic properties and identify deepfakes created with techniques like Deepfake~\cite{ntnnye18} and Face2Face~\cite{tzstn}.  Nguyen et al.~\cite{nye} proposed a model that incorporates a combination of VGG-19 and capsule networks to learn complex hierarchical representations for detecting various types of forgery, including FaceSwap~\cite{ksdt}, Facial Reenactment~\cite{tzstn}, replay attacks, and AI-generated videos. Yang et al.~\cite{yll19} proposed a model to compare 3D head poses estimated from all facial landmarks. The method considers that splicing synthesized face regions into original images can introduce errors in landmark locations. The landmark location errors can be detected by comparing the head poses estimated from the facial landmarks. Li and Lyu~\cite{ll} proposed a CNN model for deepfake detection by identifying face-warping artifacts. Their method leverages the limitations of deepfake algorithms that generate face images of lower resolutions, which results in distinctive warping artifacts when the generated images are transformed to match the original faces in the deepfake creation pipeline. By comparing the deepfake face region with surrounding pixels, resolution inconsistencies caused by face warping are identified. Li et al.~\cite{lbzycwg} proposed a technique called Face X-ray that detects the blending boundaries of images and reveals whether an input face image has been manipulated by blending two images from different sources. Sun et al.~\cite{szenqs23} proposed a virtual-anchor-based approach to robustly extract the facial trajectory, capturing displacement information. They constructed a network utilizing dual-stream spatial-temporal graph attention and a gated recurrent unit backbone to expose manipulated videos. The proposed method achieves competitive results on the FaceForensics++ dataset, demonstrating its effectiveness in detecting manipulated videos. Overall, these diverse method of detecting deepfakes based on visual cues demonstrate that subtle visual inconsistencies and artifacts can be used expose manipulated videos.

In addition to visual features, researchers have explored the use of biological signals and low-level features for deepfake detection. Y. Li~\cite{lcl18} proposed a model that combines a CNN and a recursive neural network (LRCN) to detect deepfake videos by tracking eye blinking with previous temporal knowledge. Chintha et al.~\cite{crsbwp20} proposed a modified XceptionNet architecture, which incorporates visual frames, edge maps, and dense optical flow maps alongside RGB channel data to target low-level features. The architecture isolates deepfakes at the instance and video levels, making the technique effective in detecting deepfakes. Zhao et al.~\cite{zxxdxx21} proposed a pair-wise self-consistency learning with an inconsistency image generator to train a ConvNet~\cite{lmwfds} that extracts local source features and measures their self-consistency to identify Deepfakes. 
 
D. Kim and K. Kim~\cite{kk22} proposed a facial forensic framework that uses pixel-level color features in the edge region of an image and a 3D-CNN classification model to interpret the extracted color features spatially and temporally for generalized and robust face manipulation detection. Sabir et al.~\cite{scjamn} proposed a Recurrent convolutional model, whereas~\cite{wodajo2024deepfakevideodetectionusing} proposed a Convolutional Vision Transformer (CViT) model. Y. Heo~\cite{hyk23} introduced an improved Vision Transformer model with vector-concatenated CNN feature and patch-based positioning, while~\cite{hclk} used Vision Transformer with distillation and Coccomini et al.~\cite{cmgf22} combined EfficientNet and Vision Transformers (ViT) to detect deepfakes. Hybrid CNN-Transformer approaches excel by extracting local and global features for robust and strong deepfake detection.

Despite these promising advances, current detection methods struggle to generalize across diverse datasets. Table \ref{tab:training_datasets} lists the datasets used by existing deepfake detection models for training and evaluation. They often consider only few datasets and hence generative methods for training and evaluation. 
In the next section, we introduce our approach that addresses this problem.

\begin{table}[t]
    \centering
    \begin{tabular}{l l}
        \toprule
        \textbf{Paper} & \textbf{Dataset(s) for training \& evaluation} \\
        \midrule
        Image+Video Fusion~\cite{kd21} & DFDC \\
        Selim EfficientNet B7~\cite{s} & DFDC \\
        CViT~\cite{wodajo2024deepfakevideodetectionusing} & DFDC \\
        Random cut-out~\cite{kd} & DFDC (with face cut-out augmentation) \\
        STDT~\cite{zlhwzg} & FF++, Celeb-DF\\
        ViT with distillation~\cite{hclk} & DFDC \\
        Heo et. al~\cite{hyk23} & DFDC \\
        Coccomini et. al~\cite{cmgf22} & FF++, DFDC \\
        Li et. al~\cite{lbzycwg} & FF++, CelebA-HQ (FaceShifter GAN) \\
        GenConViT (ours) & FF++, DFDC, TM, Celeb (v2), TIMIT \\
        \bottomrule
    \end{tabular}
    \caption{Datasets used in selected deepfake detection methods.}
    \label{tab:training_datasets}
\end{table}

%-------------------------------------------------------------------------
\section{Proposed Deepfake Detection Method}
\label{sec:proposed}
In this section, we propose a deepfake detection framework based on a Generative Convolutional Vision Transformer (GenConViT), which we introduce for the first time.
%We present %the used datasets (Section~\ref{sec:proposed-datasets}), the preprocessing techniques (Section~\ref{sec:proposed-preprocessing}), and % 
%our proposed Generative Convolutional Vision Transformer (GenConViT) for deepfake video detection %(Section~\ref{sec:proposed-genconvit}).
%The proposed model consists of three main steps: $1)$ video preprocessing, $2)$ feature extraction and reconstruction, and $3)$ video classification.

\begin{figure*}
  \centering
  \includegraphics[trim={15em 12em 0em 11em},clip,width=64em]{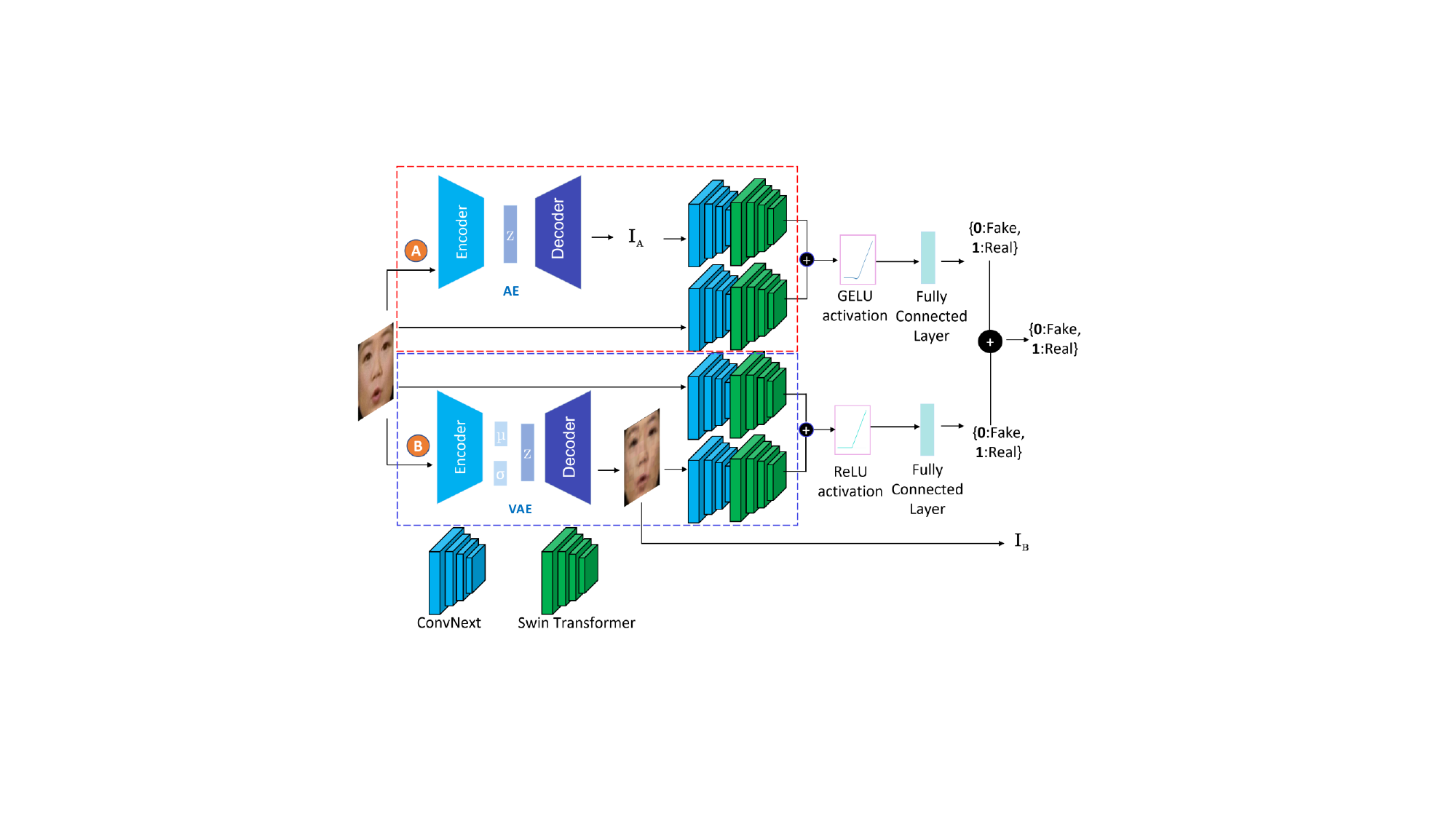}
  \caption{The Proposed GenConViT Deepfake Detection Framework.}
  \label{fig:1}
\end{figure*}

%-------------------------------------------------------------------------
%\subsection{GenConViT}
%\label{sec:proposed-genconvit}
The proposed Generative Convolutional Vision Transformer model transforms the input facial images to latent spaces, and extracts visual clues and hidden patterns within them to determine if a video is real or fake.
The proposed GenConViT model is shown in Fig.~\ref{fig:1}.
It has two independently trained networks ($A$ and $B$), and four main modules: an Autoencoder (AE), a Variational Autoencoder (VAE), a ConvNeXt layer, and a Swin Transformer. The first network ($A$) includes an AE, a ConvNeXt layer, and a Swin Transformer, while the second network ($B$) includes a VAE, a ConvNeXt layer, and a Swin Transformer. The first network uses an AE to transform images to a Latent Feature (LF) space, maximizing the class prediction probability, indicating the likelihood that a given input is deepfake.
The second network uses a VAE to maximize the probability of correct class prediction and minimize the reconstruction loss between the sample input image and the reconstructed image. Both AE and VAE models extract LFs from input facial images (extracted from video frames), which capture hidden patterns and correlations present in the learned deepfake visual artifacts. The ConvNeXt and Swin transformer models form a novel hybrid model ConvNeXt-Swin. The ConvNeXt model acts as the backbone of the hybrid model, using a CNN to extract features from the input images. The Swin Transformer, with its hierarchical feature representation and attention mechanism, further extracts the global and local features of the input. The two networks each have two ConvNeXt-Swin models, which both take in a $224 \times 224$ RGB image, as well as an LF of either the AE ($I_A$) or the VAE ($I_B$). The use of the ConvNeXt-Swin hybrid model enables the learning of relationships among the extracted LFs by the AE and VAE.

\subsection{Autoencoder and Variational Autoencoder}
An AE and a VAE consist of two networks: an Encoder and a Decoder. The Encoder of the AE maps an input image $X \in \mathbb{R}^{H \times W \times C}$ to a latent space $Z \in \mathbb{R}^{H' \times W' \times \myK}$, where $\mathsmaller{K}$ is the number of channels (features) in the output and $\mathsmaller{H'}$ and $\mathsmaller{W'}$ are the height and width of the output feature map, respectively. The Decoder of the AI maps the latent space $Z \in \mathbb{R}^{H' \times W' \times \myK}$ to an output image $X' \in \mathbb{R}^{H \times W \times C}$.
The Encoder of the AE is composed of five convolutional layers with width starting from $3$ up to $256$, with kernels of size $3 \times 3$ and a stride of $2$. Each convolutional layer is followed by ReLU non-linearity and Max-pooling of kernel size $2 \times 2$ and stride $2$. The output of the Encoder is a $256 \times 7 \times 7$ down-sampled LF. The Decoder is composed of five transposed convolutional layers with width starting from $256$ up to $3$, with kernels of size $2 \times 2$ and stride of $2$. Each transposed convolutional layer is followed by ReLU non-linearity. The output of the Decoder, $I_A$, is a reconstructed feature space of the input image with dimensions $H \times W \times C$. In this case, $I_A$ has dimensions $224 \times 224 \times 3$. The detailed configuration is shown in Table \ref{tab:ae}. 

The goal of VAE is to learn a meaningful latent representation of the input image and reconstruct the input image by performing random sampling of the latent space while minimizing the reconstruction loss.
The Encoder of VAE maps an input image $X$ to a probability distribution over a latent space $Z \in \mathbb{R}^K, Z \sim \mathcal{N}(\mu, \sigma^2)$, in which $\mu$ and $\sigma^2$ are the mean and variance of the learned distribution, respectively. 

The Encoder of the VAE is composed of four convolutional layers with width starting from $3$ up to $128$, with kernels of size $3 \times 3$ and a stride of $2$. Each convolutional layer is followed by Batch Normalization (BN) and LeakyReLU non-linearity. The output of the Encoder is a $1$-dimensional vector of length $12544$ representing the latent distributions. The Decoder is composed of four transposed convolutional layers with width starting from $256$ up to $3$, with kernels of size $2 \times 2$ and stride of $2$. Each transposed convolutional layer is followed by LeakyReLU non-linearity. The output of the Decoder, $I_B$, is a reconstructed image from the input image with dimensions $H/2 \times W/2 \times C$. In this case, $I_B$ has dimensions $112 \times 112 \times 3$. The detailed configuration is shown in Table \ref{tab:vae}. The choice of the convolutional layers for both the AE and VAE is due to the compute power and memory we had, model accuracy, extensive experiment, and training time during the training of our model.
\begin{table}
  \centering
  \caption{GenConViT model Autoencoder configuration.}
  \resizebox{\columnwidth}{!}{%
    \begin{tabular}{|l|c|c|c|c|c|c|c|c|}
    \hline
    \multicolumn{1}{|c|}{\textbf{Network}} & \multicolumn{8}{c|}{\textbf{AE Configuration}} \\
    \hline
    \textbf{Encoder} & \multicolumn{1}{c}{} & \multicolumn{5}{c|}{\textbf{Conv}} & \textbf{Kernel} & \textbf{Stride} \\
    \cline{2-9}
     & \multicolumn{2}{c|}{\text{$3$-$16$}} & \text{$16$-$32$} & \text{$32$-$64$} & \text{$64$-$128$} & \text{$128$-$256$} & \text{$3$} & \text{$1$} \\
    \hline
    \textbf{Decoder} & \multicolumn{1}{c}{} & \multicolumn{5}{c|}{\textbf{ConvTranspose}} & \multicolumn{2}{c|}{} \\
    \cline{2-9}
     & \multicolumn{2}{c|}{\text{$256$-$128$}} & \text{$128$-$64$} & \text{$64$-$32$} & \text{$32$-$16$} & \text{$16$-$3$} & \text{$2$} & \text{$1$} \\
    \hline
    \end{tabular}%
  }
  \label{tab:ae}%
\end{table}%
\begin{table}
  \centering
  \caption{GenConViT model Variational Autoencoder configuration.}
  \resizebox{\columnwidth}{!}{%
    \begin{tabular}{|l|c|c|c|c|c|c|c|}
    \hline
    \textbf{Network} & \multicolumn{7}{c|}{\textbf{VAE Configuration}} \\
    \hline
    \textbf{Encoder} & \multicolumn{2}{c|}{} & \multicolumn{3}{c|}{\textbf{Conv}} & \textbf{Kernel} & \textbf{Stride} \\
    \cline{2-8}
     & \multicolumn{2}{c|}{\text{$3$-$16$}} & \text{$16$-$32$} & \text{$32$-$64$} & \text{$64$-$128$} & \text{$3$} & \text{$2$} \\
    \hline
    \textbf{Decoder} & \multicolumn{1}{c}{} & \multicolumn{4}{c|}{\textbf{ConvTranspose}} & \multicolumn{2}{c|}{} \\
    \cline{2-8}
     & \multicolumn{2}{c|}{\text{$256$-$64$}} & \text{$64$-$32$} & \text{$32$-$16$} & \text{$16$-$3$} & \text{$2$} & \text{$2$} \\
    \hline
    \end{tabular}%
  }
  \label{tab:vae}%
\end{table}%

\subsection{ConvNeXt-Swin Hybrid}
The ConvNeXt-Swin Transformer architecture is a hybrid CNN-Transformer model that combines the strengths of ~\cite{lmwfds} and Swin Transformer~\cite{llchwzlg} architectures for deepfake detection task. The ConvNeXt model is a CNN architecture that has shown impressive performance in the image recognition tasks, by extracting high-level features from images through a series of convolutional layers. The Swin Transformer is a transformer-based model that uses a self-attention mechanism to extract both local and global features.

The GenConViT model leverages the strengths of both architectures by using ConvNeXt as the backbone for feature extraction and the Swin Transformer for feature processing. In our proposed method, the ConvNeXt architecture extracts high-level features from images, which are then passed through a HybridEmbed module to embed the features into a compact and informative vector. The resulting vector is then passed to the Swin Transformer model. The ConvNeXt backbone consists of multiple convolutional layers that extract high-level features from input images and the LFs from AE or VAE. We use pre-trained ConvNeXt and Swin Transformer models, which are trained on an ImageNet dataset. 

After extracting learnable features by the ConvNeXt backbone, we pass the feature maps through a HybridEmbed module. The HybridEmbed module is designed to extract feature maps from the ConvNeXt, flatten them, and project them to an embedding dimension of $768$. It consists of a $1 \times 1$ convolutional layer, which takes the feature maps from the backbone and reduces their channel dimension to the desired embedding dimension. The resulting feature maps are flattened and transposed to obtain a sequence of feature vectors, which are then further processed by the Swin Transformer. 

The GenConViT’s network $A$ consists of two Hybrid ConvNeXt-Swin models that take in an LF of size $224 \times 224 \times 3$ generated by the AE ($I_A$) and an input image of the same size. The models output a feature space of size $1,000$, which is then concatenated. A linear mapping layer of $2$ then transforms this combined feature vector into a class prediction, corresponding to the probabilities for the real and fake classes.
Network $B$ has the same configuration as $A$, but it uses VAE and outputs both class prediction probability and the reconstructed image of size $224 \times 224 \times 3$.
Finally, the prediction of network A and network B are averaged to obtain the final real/fake prediction. 

In summary, our proposed GenConViT method introduces a hybrid architecture for deepfake detection, combing the feature extraction capabilities of the pre-trained ConvNeXt-Swin architectures and generative components using AE and VAE. This setup helps our model to capture subtle inconsistencies in deepfake videos. The open-source code for GenConViT is available at \url{https://github.com/erprogs/GenConViT}.

%----------------------------------------------------------------
\section{Evaluation}
\label{sec:evaluation}
We conducted extensive experiments on various configurations of AE and VAE, as well as different variants of CNN and Transformer models. Our findings suggest that a hybrid architecture using ConvNeXt and the Swin Transformer performs well. 
We first describe the experimental setup in Section~\ref{sec:evaluation-setup}, after which the results are presented and discussed in Section~\ref{sec:evaluation-results}.

\subsection{Experimental Setup}
\label{sec:evaluation-setup}

To assess GenConViT’s performance, we used multiple evaluation metrics, including classification accuracy, F$1$ score, Receiver Operating Characteristic (ROC) curve, and Area Under the Curve (AUC) value.

The implementation details, datasets, and preprocessing methodology is described in Section~\ref{sec:evaluation-setup-implementation}, \ref{sec:evaluation-setup-datasets}, and \ref{sec:evaluation-setup-preprocessing}, respectively.

\subsubsection{Implementation Details}
\label{sec:evaluation-setup-implementation}
\begin{figure}
    \centering
    \begin{subfigure}{0.227\textwidth}
        \centering
        \includegraphics[width=\linewidth]{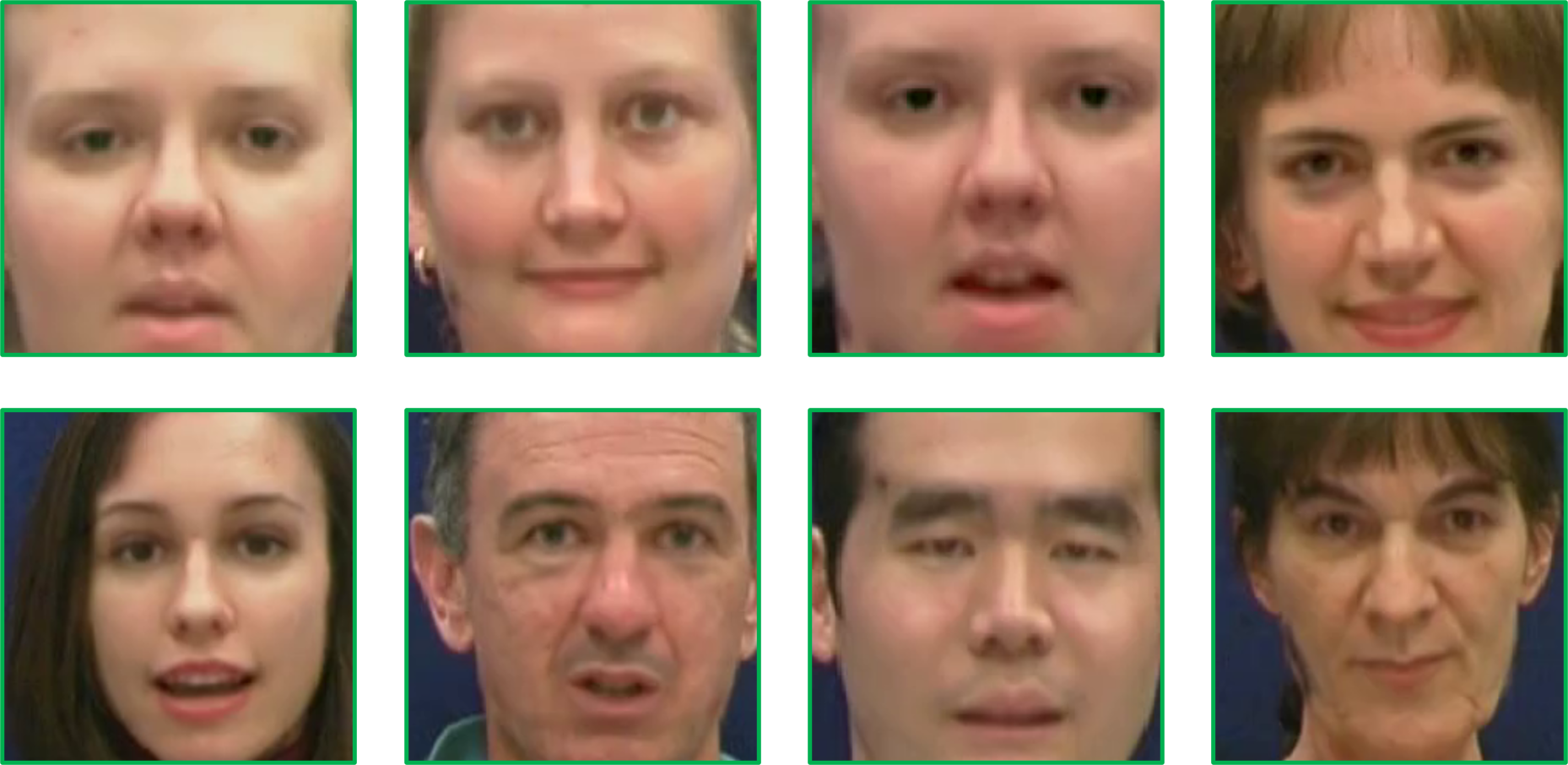}
        \caption{Input Sample Images}
        \label{subfig:input_images}
    \end{subfigure}
    \begin{subfigure}{0.227\textwidth}
        \centering
        \includegraphics[width=\linewidth]{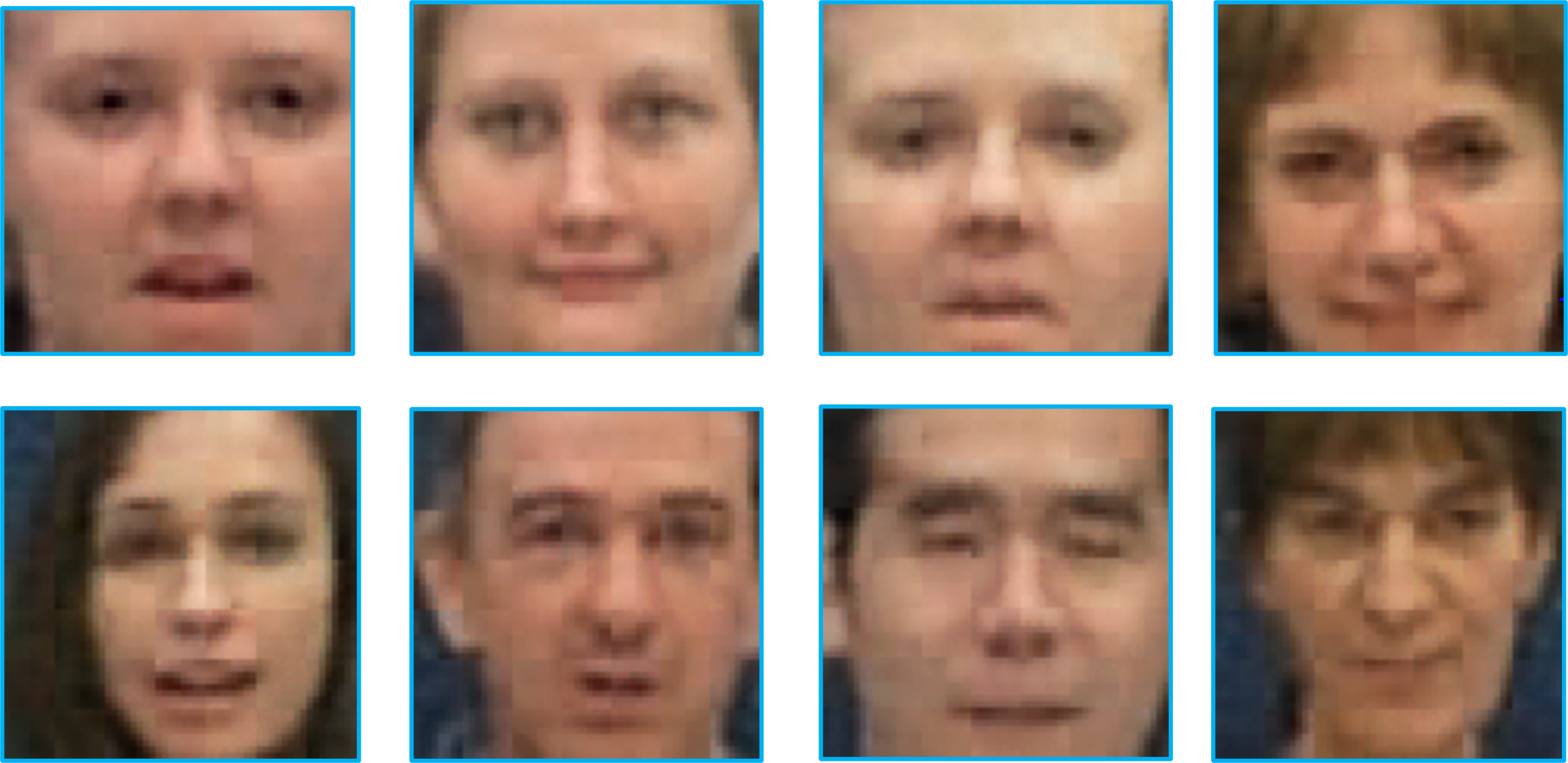}
        \caption{Images Generated by $B$}
        \label{subfig:reconstructed_images}
    \end{subfigure}
    \caption{Generated Images ($I_B$) from Input Samples (a) using Network $B$ (b)}
    \label{fig:2}
\end{figure}

The network $A$ and $B$ were trained to classify real and fake videos, while $B$ was additionally trained to the reconstructed images, of which some examples are shown in Fig.~\ref{fig:2}. 
Therefore, Network $A$ is trained using the Cross Entropy loss, while Network $B$ is trained using the Cross Entropy and the MSE loss. We used the \emph{timm}~\cite{wp19} library to load the class definitions and the weight of the pretrained ConvNext and Swin Transformer. Due to our limited resources and large training dataset, we implemented the "tiny" model versions of both architectures, namely $\texttt{convnext\_tiny}$ and $\texttt{swin\_tiny\_patch4\_window7\_224}$, both trained on ImageNet-$1$k.

Both network, $A$ and $B$, were trained using the Adam optimizer with a learning rate of $0.0001$ and weight decay of $0.0001$. The Albumentation~\cite{bikpdk20} library was used for data augmentation, and the following augmentation techniques were used with a strong augmentation rate of $90\%$: RandomRotate, Transpose, HorizontalFlip, VerticalFlip, GaussNoise, ShiftScaleRotate, CLAHE, Sharpen, IAAEmboss, RandomBrightnessContrast, and HueSaturationValue. The training data was normalized. The batch size for network $A$ is set to $32$ and for network $B$, it was set to $16$. Both networks were trained for $30$ epochs. 

\subsubsection{Datasets}
\label{sec:evaluation-setup-datasets}
In our work, we utilized five datasets to train, validate, and test our model: DFDC~\cite{dhpbf,dbplhwf}, TrustedMedia (TM)~\cite{ccwn22}, DeepfakeTIMIT (TIMIT)~\cite{km18,sl09}, Celeb-DF (v$2$)~\cite{lysql20,Li2022}, and FaceForensics++ (FF++)~\cite{rcvrtn}. DFDC and FF++ are well-known benchmark datasets for deepfake detection. TM is created using a diverse range of deepfake manipulation techniques. 

The DFDC dataset is the largest publicly available dataset and contains over $100,000$ high-resolution real and fake videos. The dataset is created using $3,426$ volunteers, and the videos are captured in various natural settings, different angles, and lighting conditions. The dataset is created using eight deepfake creation techniques.

The FF++ dataset comprises $1,000$ original videos collected from YouTube, which have been manipulated using four automated face manipulation methods: Deepfakes, Face2Face, FaceSwap, and NeuralTextures~\cite{tzn}. The dataset includes compressed videos with quantization parameters of $c23$ and $c40$ in various video resolutions. The TM dataset consists of $4,380$ fake and $2,563$ real videos, with multiple video and audio manipulation techniques. The TM dataset is used only in the training phase. The Celeb-DF (v2) dataset consists of $890$ real videos and $5,639$ videos deepfake videos. 

We randomly extracted approximately $30$ frames per video from each dataset to ensure diversity in our training data. To mitigate the ratio between fake and real videos in DFDC and TM datasets, we extracted a higher number of frames from the real videos. The DFDC dataset has a ratio of $6:1$ for fake to real videos, and TM has approximately $2:1$ ratio.

By using a variety of datasets, utilizing multiple deepfake creation techniques, and captured in multiple settings, we aim to provide better generalization and robustness to varying environments. Notably, our dataset is trained (and evaluated) using significantly more datasets (see Table~\ref{tab:training_datasets}). 

%------------------------------------------------------------------------
\subsubsection{Video Preprocessing}
\label{sec:evaluation-setup-preprocessing}
We preprocess the frames of the videos of the datasets such that we can work with images that only contain information about the faces, and are correctly labeled.
The preprocessing component in DL plays a critical role in preparing raw datasets for training, validation, and testing. The proposed model focuses on the face region, which is crucial in deepfake generation and synthesis mechanisms. We therefore preprocess the videos using a series of image processing operations. These operations include the following steps:
\begin{enumerate}
\item Extracting the face region from each videos using OpenCV, face\_recognition~\cite{g}, and BlazeFace~\cite{bkvrg} face recognition deep learning libraries;
\item Resizing the input (facial) image to a $224 \times 224$ RGB format, where the dimensions of the input image are $H \times W \times C$ with $H=224$ representing the height, $W=224$ representing the width, and $C=3$ representing the RGB channels;
\item Verifying extracted face region images quality manually.
\end{enumerate}

After the face regions were extracted, we manually review them to fix two problems. As noted in~\cite{dbplhwf}, (1) deepfake videos may contain pristine frames within them, and (2) face regions may not always be accurately detected by the face recognition frameworks used to extract them. To address this issue, we manually reviewed the images and excluded images that did not contain a face, or were deemed to be real image within the fake class. This approach allowed us to curate a fake class dataset comprising only relevant and potentially manipulated face images.
 
As a result from applying the preprocessing steps on the datasets, we collected a total of $1,004,810$ images. To train, validate, and test our model, the images were divided into a ratio of approximately $80:15:5$, resulting in $826,756$ images for training, $130,948$ images for validation, and $47,106$ images for testing. Note that this preprocessed test set was only used for internal evaluation, the main experiments of this paper was performed on videos (in contrast to individual frames). Namely, for evaluation, we held out $3,972$ videos from the DFDC, DeepfakeTIMIT, Celeb-DF (v$2$), and FF++ datasets for testing.
To have a single prediction per video, we extracted $15$ frames from each video and averaged the corresponding resulting predictions.

%-------------------------------------------------------------------------
\subsection{Experimental Results \& Discussion}
\label{sec:evaluation-results}
In this section, we present the experimental results and discuss the performance of our proposed GenConViT model.

Table~\ref{tab:result_all} summarizes the accuracy results of the proposed GenConViT model, as well as its internal network A and B, on the evaluation datasets. The table also separately shows the accuracy for only the real and only the fake samples, to show potential discrepancies in performance. Note that the TIMIT dataset does not contain any real videos. 

The results demonstrate that GenConViT delivers strong performance across various datasets.
The individual networks A and B also both demonstrate decent performance, but are outperformed by their combination.
Additionally, the measured performance is similar for both the real and fake videos, except for Celeb-DF (v2), which has a worse accuracy for detecting real videos compared to fake videos.

\begin{table*}[h]
\centering
\caption{Comparison of accuracy values (\%) of GenConViT and its internal networks across multiple evaluation datasets.}
\begin{tabular}{r|ccc|ccc|ccc}
\toprule
\multirow{2}{*}{Dataset} 
  & \multicolumn{3}{c|}{GenConViT} 
  & \multicolumn{3}{c|}{GenConViT A} 
  & \multicolumn{3}{c}{GenConViT B} \\
  & ALL & REAL & FAKE
  & ALL & REAL & FAKE
  & ALL & REAL & FAKE \\
\midrule
DFDC   &  98.50      &  98.70     &  98.45       & 97.50       &  98.70     & 97.20      &  98.45     &  98.70       &  95.52     \\
FF++   &  97.00     &  95.58       &  98.50     &  95.57     &  94.12     &  95.56     &  96.80     & 95.58       &  98.02    \\
TIMIT  &  98.28     &  -     &  98.28     &  97.65     &  -     & 97.50      &  97.81     &  -     &  97.80     \\
Celeb-DF (v2) &  90.94     &  83.00     &  98.80     &  85.42     &  70.22     &  93.38     &  83.97     &  55.00       & 99.38     \\
\midrule
\textit{Average} &  96.05     &  92.42      &  98.50      &  94.03      &  87.68      &  95.91      &  94.26      &  83.09        & 97.68     \\
\bottomrule
\end{tabular}
\label{tab:result_all}
\end{table*}

\begin{figure}[tbp]
  \centering
  \includegraphics[width=\linewidth]{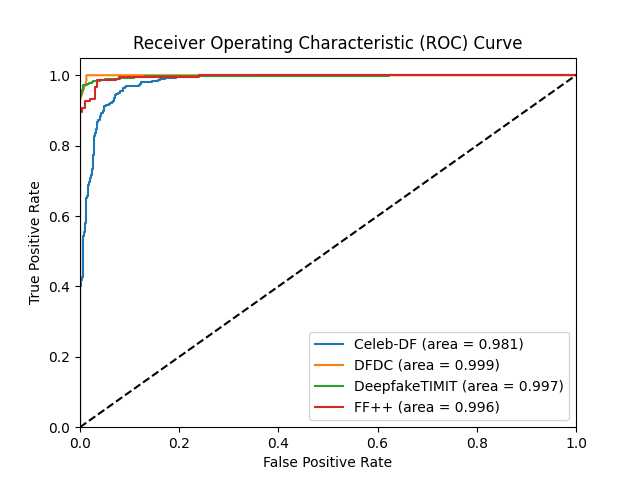}
  \caption{ROC curve illustrating the model's discrimination ability between real and fake classes.}
  \label{fig:3}
\end{figure}

For completeness, Table \ref{tab:genconvitauc} and Table~\ref{tab:genconvitf1} present the AUC value and F1-score, respectively. To further investigate GenConViT performance, Figure ~\ref{fig:3} presents the resulting ROC curve. Both networks A and B have relatively similar results.

\begin{table}[h]
  \centering
  \caption{AUC values (\%) for GenConViT, GenConViT($A$) and GenConViT($B$) in each dataset.}
    \begin{tabular}{r|lll}
    \toprule
    Dataset & GenConViT & GenConViT($A$) & GenConViT($B$) \\
    \midrule
    DFDC  & $99.9$ & $99.9$  & $99.9$ \\
    FF++ & $99.6$ & $99.1$ & $99.6$\\
    Celeb-DF (v$2$) & $98.1$ & $97.8$  & $94.1$ \\
    \bottomrule
    \end{tabular}%
  \label{tab:genconvitauc}%
\end{table}%

\begin{table}[h]
  \centering
  \caption{F$1$ scores (\%) for GenConViT, GenConViT($A$) and GenConViT($B$) in each dataset.}
    \begin{tabular}{r|lll}
    \toprule
    Dataset & GenConViT & GenConViT($A$) & GenConViT($B$) \\
    \midrule
    DFDC  & $99.1$ & $98.4$  & $98.4$ \\
    FF++  & $95.5$ & $94.9$  & $96.8$ \\
    TIMIT  & $98.3$ & $97.5$  & $97.8$ \\
    Celeb-DF (v$2$) & $91.6$ & $95.2$  & $89.0$ \\
    \bottomrule
    \end{tabular}%
  \label{tab:genconvitf1}%
\end{table}%

Overall, the proposed GenConViT model has an average accuracy of $95.8\%$ and an AUC value of $99.3\%$ across the tested datasets. These results highlight our model's robust performance in detecting deepfake videos, demonstrating its potential for practical applications in the field.

\section{Comparison with State of the Art}
\label{sec:evaluation-results-sota}
We compared our GenConViT model with several other state-of-the-art methods on the DFDC dataset (Table~\ref{tab:sotadfdc}), the FF+ dataset (Table~\ref{tab:sotaff}), and the three subsets of FF++ (Table~\ref{tab:sotaffsubset}). Note that the testing sub-datasets of DFDC and FF++ are usually not shared in the scientific literature. Hence, the exact evaluated datasets may differ, although comparing the reported values gives an idea of how their performances compare. Table \ref{tab:training_datasets} lists the datasets used by other deepfake detection models for training and evaluation. Note again that Table~\ref{tab:training_datasets} showed that our dataset is trained and evaluated using the most datasets. Other datasets do not evaluate against out-of-domain deepfakes, and hence may not actively expose their potential limitations. 

From Table~\ref{tab:sotadfdc} (DFDC dataset), we can observe that previous approaches achieved accuracy values ranging from $91.5\%$ to $98.24\%$ on the DFDC dataset. Notable models included Khan~\cite{kd21} ($91.69\%$), and Thing~\cite{t} ($92.02\%$), Selim~\cite{s} ($97.2\%$). Some models excelled in additional metrics, such as the STDT~\cite{zlhwzg} model with $97.44\%$ accuracy, $99.1\%$ AUC, and $98.48\%$ F$1$-score. In comparison to previous approaches, our proposed model demonstrated excellent performance, achieving an average accuracy of $98.5\%$, an AUC of $99.9\%$, and an F$1$-score of $99.1\%$.

\begin{table}[h]
  \centering
  \caption{GenConViT model evaluation metrics on the \textbf{DFDC} dataset compared with other state-of-the-art models.}
  \begin{tabular}{r|lll}
    \toprule
    Model & Accuracy (\%) & AUC (\%) & F$1$-Score (\%) \\
    \midrule
    Image+Video Fusion~\cite{kd21} & $91.69$ & - & - \\
    Selim EfficientNet B7~\cite{s} & $97.20$ & $90.60$ & - \\
    CViT~\cite{wodajo2024deepfakevideodetectionusing} & $91.50$ & $91.00$ & - \\
    Thing~\cite{t} & $92.02$ & $97.61$ & - \\
    Random cut-out~\cite{kd} & $98.24$ & - & - \\
    STDT~\cite{zlhwzg} & $97.44$ & $99.10$ & $98.48$ \\
    ViT with distillation~\cite{hclk} & - & $97.80$ & $91.90$ \\
    Heo et. al~\cite{hyk23} & - & $97.80$ & $91.90$ \\
    Coccomini et. al~\cite{cmgf22} & - & $95.10$ & $88.00$ \\
    GenConViT (ours) & $\textbf{98.50}$ & $\textbf{99.90}$ & $\textbf{99.10}$ \\
    \bottomrule
  \end{tabular}%
  \label{tab:sotadfdc}%
\end{table}%

\begin{table}[h]
  \centering
  \caption{GenConViT model accuracy and AUC on the FF++ dataset compared to with other state-of-the-art models.}
    \begin{tabular}{r|lll}
    \toprule
    Model & Accuracy (\%) & AUC (\%) & F1-Score (\%) \\
    \midrule
    Li et. al~\cite{lbzycwg} & $\textbf{97.73}$ & $98.52$ & - \\
    Image+Video Fusion~\cite{kd21} & $99.52$ & $99.64$ & $\textbf{99.28}$ \\
    Random cut-out~\cite{kd} & $97.00$ & $99.28$ & - \\
    GenConViT (ours) & $97.00$ & $\textbf{99.60}$ & $97.1$ \\
    \bottomrule
    \end{tabular}%
  \label{tab:sotaff}%
\end{table}%

\begin{table}[h]
  \centering
  \caption{GenConViT model accuracy (\%) on the three subsets of FF++ dataset compared to with other state-of-the-art models.}
    \begin{tabular}{r|lll}	
    \toprule
    Model & Deepfakes & Face2Face & NeuralTextures \\
    \midrule
    Li et. al~\cite{lbzycwg} & $\textbf{$99.17$}$ & $97.73$ & - \\
    Coccomini~\cite{cmgf22} & $87$  & - & $69$ \\
    Random cut-out~\cite{kd}  & $98.57$ & $98.57$ & $90.71$ \\
    GenConViT (ours)  & $92.27$ & $\textbf{98.00}$ & $\textbf{97.00}$ \\
    \bottomrule
    \end{tabular}%
  \label{tab:sotaffsubset}%
\end{table}%

\section{Ablation Study}
\label{sec:ablation}
To evaluate the generalization capability of our proposed model, we conducted an out-of-distribution (OOD) ablation study. We trained our model on four of the five available datasets, holding out the fifth dataset entirely (unseen during training and validation). We repeated this procedure for different held-out datasets and systematically varied the hyperparameters (such as the layers of the CNN depth, width, and learning rates) to select the best performing model variant. Moreover, to optimize computation, we performed training for ablation experiments on a randomly selected subset of approximately 5\% of the images from the training datasets. 
 
Table \ref{tab:ablation} summarizes the performance of the model across
various scenarios. In each scenario, the column labeled “Held-
out” denotes the dataset excluded from training, whereas the column labeled “Test set” denotes the evaluation dataset. Then, the other columns presents the performance on the test set when training using all test datasets, except for the held-out dataset. Cells highlighted by bold text represent significant drops in performance due to holding out the corresponding dataset during training.

We found that our model struggled to detect fake videos in the held-out dataset in an OOD setting, indicating that it still faces challenges in generalizing to unseen, more hyper-realistic deepfake images. 
Notably, when the model encounters a fully unseen,
hyper-realistic dataset, it struggles to detect
fake videos, resulting in a substantial drop in accuracy on the
fake class. For instance, in the Celebv2 held-out scenario,
the model’s overall accuracy on Celebv2 can be as low as
55.67\%, with 11.56\% accuracy on the fake samples.  These
results suggest that, although our model performs well on in-
domain data (TM, DeepfakeTIMIT, DFDC, and FF++), it faces challenges in generalizing to significantly different or higher-fidelity deepfakes.

These experiments suggest that deeper or wider CNN architectures alone do not necessarily guarantee improved robustness to domain shifts. Despite experimenting with various hyperparameters (e.g., layer depth, width, and learning rates), the drop in fake accuracy remained significant when the held-out dataset was substantially different from the training sets.

Another noteworthy observation is that certain datasets, like DeepfakeTIMIT, appear easier to generalize to. This may be due to their relatively simpler manipulations. In contrast, datasets such as Celebv2 or DFDC contain higher-quality forgeries and more varied manipulations, creating a more challenging OOD scenario. Consequently, our study underscores the importance of curating diverse, high-fidelity training sets when aiming to build robust deepfake detection models. Therefore, our proposed GenConViT model was trained on 5 datasets, representing a large variety of deepfakes and settings.

Overall, these findings confirm that while our model
achieves promising performance on in-domain data, it remains
sensitive to domain shifts, particularly when encountering
previously unseen or more realistic deepfake manipulations.
This remains an area for future work to explore.

\begin{table}[h]
\centering
\caption{Ablation Results Across All Training Scenarios (40k samples)}
\begin{tabular}{ll|cccccc}
\toprule
\textbf{Held-out} & \textbf{Test set} & \textbf{Acc. (\%)} 
& \textbf{Real Acc. (\%)} & \textbf{Fake Acc. (\%)} \\
\midrule

%---------------------------------------------
% SCENARIO 2
\multirow{5}{*}{\shortstack{Celebv2}} 
& TM              & 88.30 & 87.38 & 89.19  \\
& TIMIT   & 99.96 & 99.91 & 100.00 \\
& DFDC            & 89.11 & 89.11 & 89.11 \\
& FF++   & 99.56 & 99.71 & 95.24 \\
& \textbf{Celebv2}         & \textbf{55.67} & 99.78 & \textbf{11.56} \\
\midrule

%---------------------------------------------
% SCENARIO 3
\multirow{4}{*}{\shortstack{DFDC}} 
& TM              & 88.30 & 87.38 & 89.19 \\
& TIMIT   & 99.42 & 98.72 & 100.00 \\
& FF++   & 99.53 & 99.64 & 96.53 \\
& Celebv2         & 97.83 & 97.11 & 98.56 \\
& \textbf{DFDC}         & \textbf{62.08} & 95.31 & \textbf{28.84} \\
\midrule

%---------------------------------------------
% SCENARIO 4
\multirow{5}{*}{\shortstack{FF++}} 
& TM              & 88.88 & 89.76 & 88.04 \\
& TIMIT   & 99.61 & 99.15 & 100.00 \\
& Celebv2         & 96.92 & 95.67 & 98.18 \\
& DFDC            & 88.53 & 91.29 & 85.78 \\
& \textbf{FF++}   & \textbf{64.86} & 86.78 & \textbf{44.20} \\
\midrule

%---------------------------------------------
% SCENARIO 5
\multirow{5}{*}{\shortstack{TIMIT}} 
& TM              & 88.38 & 88.81 & 87.97 \\
& Celebv2         & 96.77 & 94.49 & 99.04 \\
& DFDC            & 88.66 & 99.36 & 87.96 \\
& FF++   & 99.51 & 99.68 & 94.59 \\
& \textbf{TIMIT}  & 93.44 & 95.98 & 91.36 \\
\midrule

%---------------------------------------------
% SCENARIO 6
\multirow{5}{*}{\shortstack{TM}} 
& TIMIT   & 99.88 & 99.74 & 100.00 \\
& DFDC            & 88.76 & 91.93 & 85.58 \\
& FF++   & 99.67 & 99.84 & 94.72 \\
& Celebv2         & 97.73 & 98.53 & 96.93 \\
& \textbf{TM} & \textbf{76.80} & 97.78 & \textbf{56.66} \\
\bottomrule
\end{tabular}
\label{tab:ablation}
\end{table}
%-------------------------------------------------------------------------

\section{Conclusion}
\label{sec:conclusion}
In this work, we proposed a Generative Convolutional Vision Transformer (GenConViT) that extracts visual artifacts and latent data distributions to detect deepfake videos. GenConViT combines the ConvNext and  Swin transformer architectures to learn from local and global image features of a video, as well as an AE and VAE to learn from internal data representations. Our approach provides an effective solution for identifying fake videos while preserving media integrity. Through extensive experiments on a diverse dataset, including DFDC, FF++, DeepfakeTIMIT, and Celeb-DF (v$2$), our GenConViT model demonstrated improved and robust performance with high classification accuracy, F$1$ scores, and AUC values. Our ablation study reveals challenges in generalizing to unseen or more complex manipulations, highlighting the need for the further research to improve domain adaptability. Overall, our proposed GenConViT model offers a promising approach for accurate and reliable deepfake video detection.
As our model is open source, its practical use was already demonstrated: it was used by TrueMedia.org, a non-profit organisation to detect deepfakes and support fact-checking efforts~\cite{tm2025}.

\section*{Acknowledgment}
\thanks{This research was supported by Addis Ababa University Research Grant for the Adaptive Problem-Solving Research. Reference number RD/PY-183/2021. Grant number AR/048/2021, and the Research Foundation – Flanders (FWO under project grant G0A2523N), the Flemish government (COM-PRESS project, within the relanceplan Vlaamse Veerkracht), IDLab (Ghent University – imec), Flanders Innovation and Entrep-reneurship (VLAIO), and the European Union.}

{\small
\bibliographystyle{ieee_fullname}
\bibliography{ref}
}

\end{document}